\documentclass{article}

\usepackage{arxiv}

\usepackage[utf8]{inputenc} % allow utf-8 input
\usepackage[T1]{fontenc}    % use 8-bit T1 fonts
\usepackage{hyperref}       % hyperlinks
\usepackage{url}            % simple URL typesetting
\usepackage{booktabs}       % professional-quality tables
\usepackage{amsfonts}       % blackboard math symbols
\usepackage{nicefrac}       % compact symbols for 1/2, etc.
\usepackage{microtype}      % microtypography 
\usepackage{graphicx}
\usepackage{natbib}
\usepackage{doi}
\usepackage{float}
\usepackage{multirow} 
\usepackage{amsmath}
\usepackage{stmaryrd}
\usepackage{amssymb} 
\usepackage{subcaption}
\usepackage{comment}

\title{High Noise Scheduling is a Must}

\author{ \href{https://orcid.org/0000-0000-0000-0000}{\includegraphics[scale=0.06]{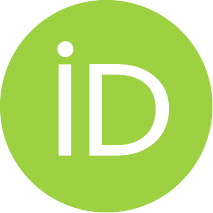}\hspace{1mm}Mahmut S. Gokmen} \\
	Department of Computer Science\\
	University of Kentucky\\ 
	\texttt{mselmangokmen@uky.edu} \\
	\And
	\href{https://orcid.org/0000-0000-0000-0000}{\includegraphics[scale=0.06]{orcid.pdf}\hspace{1mm}Cody Bumgardner} \\
	Department of Internal Medicine\\
        Institute for Biomedical Informatics\\
	University of Kentucky\\  
	\texttt{cody@uky.edu} \\
  \And
	\href{https://orcid.org/0000-0000-0000-0000}{\includegraphics[scale=0.06]{orcid.pdf}\hspace{1mm}Jie Zhang} \\
	Department of Radiology\\
	University of Kentucky\\  
	\texttt{jie.zhang1@uky.edu} \\
 \And
 \href{https://orcid.org/0000-0000-0000-0000}{\includegraphics[scale=0.06]{orcid.pdf}\hspace{1mm}Ge Wang} \\
        Department of Biomedical Engineering\\
	Rensselaer Polytechnic Institute\\  
	\texttt{wangg6@rpi.edu} \\
 \And
	\href{https://orcid.org/0000-0000-0000-0000}{\includegraphics[scale=0.06]{orcid.pdf}\hspace{1mm}Jin Chen} \\
	Department of Medicine\\
        Department of Biomedical Informatics and Data Science\\
	University of Alabama at Birmingham\\  
	\texttt{jinchen@uab.edu} \\
}
\renewcommand{\shorttitle}{High Noise Scheduling Is A Must}

\begin{document}
\maketitle

\begin{abstract} 
Consistency models possess high capabilities for image generation, advancing sampling steps to a single step through their advanced techniques. Current advancements move one step forward consistency training techniques and eliminates the limitation of distillation training. Even though the proposed curriculum and noise scheduling in improved training techniques yield better results than basic consistency models, it lacks well balanced noise distribution and its consistency between curriculum. In this study, it is investigated the balance between high and low noise levels in noise distribution and offered polynomial noise distribution to maintain the stability. This proposed polynomial noise distribution is also supported with a predefined Karras noises to prevent unique noise levels arises with Karras noise generation algorithm. Furthermore, by elimination of learned noisy steps with a curriculum based on sinusoidal function increase the performance of the model in denoising. To make a fair comparison with the latest released consistency model training techniques, experiments are conducted with same hyper-parameters except curriculum and noise distribution. The models utilized during experiments are determined with low depth to prove the robustness of our proposed technique. The results show that the polynomial noise distribution outperforms the model trained with log-normal noise distribution, yielding a 33.54 FID score after 100,000 training steps with constant discretization steps. Additionally, the implementation of a sinusoidal-based curriculum enhances denoising performance, resulting in a FID score of 30.48. 
\end{abstract}

% keywords can be removed

\section{Introduction}
Consistency models are the recently released family of generative models. They are fast  evolving based knowledge distillation training techniques. Unlike other generative models such as score-based diffusion models~\cite{Song2020-sb}, it is not required to numerous samplings to generate high quality samples since consistency models can generate it in a single step. Also, consistency models preserves flexibility and computation theory for generating samples in different number of inference steps. 

Consistency models uses two different training technique for training which are represented as consistency distillation (CD) and consistency training (CT)~\cite{Song2023-cm}. The CD training requires a pre-trained model for distilling the knowledge into consistency model. The CT training let the model learn directly from data. Although recent studies shows that CD is better than CT, CD requires high computational source for distilling data from pre-trained models. Additionally, the best outperforming results are limited by the distilled pre-trained models in CD.  

To overcome the drawbacks of CD and advance CT further, \cite{Song2023-imp} proves that CT outperforms CD by eliminating Exponential Moving Average (EMA) for the teacher network and improving curriculum with a step-wise increase relied on the current training steps. Furthermore, noise scheduling is also improved by adopting the log-normal distribution to sample noise levels. 

Primarily, the Log-normal noise scheduling tends to introduce high-weighted low-level noises which are accumulated at the around of standard deviation of Karras noises generated according to discretization steps. Although this approach is able to sustain to introduce high-weighted low-level noises in noise scheduling, that dramatically decreases the variation on high level noises. In this study, we introduce that high level noises is a must in the noise scheduling to meet the basic rules for consistency models. This assumption is grounded in the main principle of consistency models, which leverage learning the denoising process from $x_{(t+1)}$, handled by $\xi$, to $x_{(t)}$, handled by $\theta$, where $\xi$ and $\theta$ represent the index of noise in discretization steps and the learnable parameters of the model, respectively. Therefore, it is crucial to incorporate a high variety of noise levels during training to ensure effective denoising from $x_{(t+1)}$ to $x_{(t)}$ \cite{Song2023-imp}. In a situation the model rarely experienced the high noise levels decreases the denoising performance.

\begin{figure}[ht]
    \centering
    \begin{subfigure}{0.45\textwidth}
        \centering
        \includegraphics[width=\textwidth]{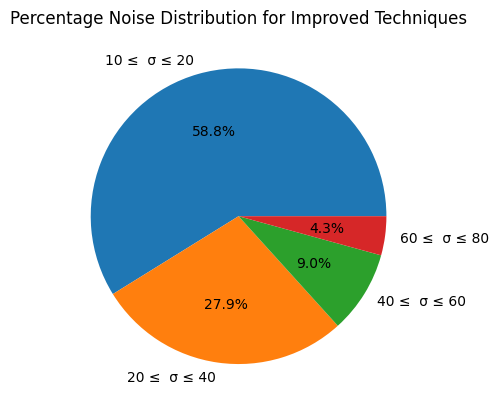}
        \caption{Noise distribution pie chart for improved technique}
        \label{fig:fig_pie_a}
    \end{subfigure}
    \begin{subfigure}{0.45\textwidth}
        \centering
        \includegraphics[width=\textwidth]{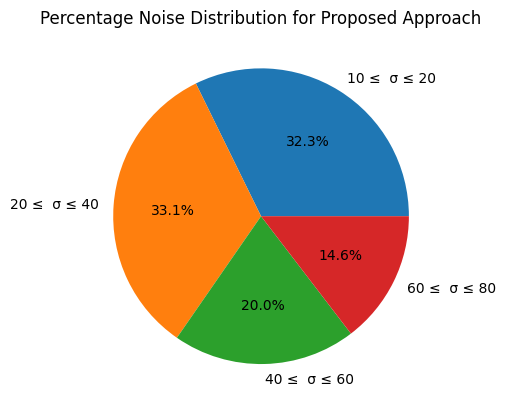}
        \caption{Noise distribution pie chart for polynomial technique}
        \label{fig:fig_pie_b}
    \end{subfigure}
    \caption{Percentage noise distribution for improved techniques (a) and percentage noise distribution for polynomial noise scheduling (b) }
    \label{fig:fig_pie}
\end{figure}

To highlight the disparities between noise scheduling techniques, including improved methods and our proposed polynomial approach, Figure \ref{fig:fig_pie} illustrates noise distributions using pie chart slices divided into four sections: noise values between $10 \leq \sigma \leq 20$, $20 \leq \sigma \leq 40$, $40 \leq \sigma \leq 60$, and $60 \leq \sigma \leq 80$. The reason for setting the minimum noise level bound in the pie chart is to enhance the visibility of the other ratios. This decision is based on the fact that noise levels between $0$ and $10$ exhibit higher dominance compared to the higher noise levels.

\iffalse
To prove this assumption, a histogram for noise distribution which relies on log-normal noise distribution is depicted in Figure \ref{fig:fig_hist_a}.

\begin{figure}[ht]
    \centering
    \begin{subfigure}{0.45\textwidth}
        \centering
        \includegraphics[width=\textwidth]{figures/noise_dist_imp.png}
        \caption{Noise distribution histogram for improved technique}
        \label{fig:fig_hist_a}
    \end{subfigure}
    \begin{subfigure}{0.45\textwidth}
        \centering
        \includegraphics[width=\textwidth]{figures/noise_dist_our.png}
        \caption{Noise distribution histogram for polynomial technique}
        \label{fig:fig_hist_b}
    \end{subfigure}
    \caption{Distribution of noise levels across mini-batches for improved techniques (a) and polynomial noise scheduling (b), replace this histogram with a pie chart that shows distribution of high and low noise }
    \label{fig:fig_hist}
\end{figure}
\fi

Additionally, noise scheduling crucially depends on curriculum to provide high variety of noise levels on mini-batches. The marginal changes on the discretization steps loads high variety of noises noises which are never experienced before by the model. Even Karras noise scheduling algorithm generates close noise levels according to the discretization steps, it is observable that noise levels differ each other. Thus, we claim that the marginal changes on curriculum creates an unstability on learning parameters and output of losses \cite{Song2023-imp}.

To overcome we propose three modification on noise scheduling and curriculum; 
\begin{itemize}
  \item Experiments investigating the advantages of noise scheduling, which includes high noise levels, and proposing a novel noise scheduling method based on a polynomial function.
  \item Creating a predefined Karras noise vector $\sigma_{s_1}$ to prevent production of unique noise levels to maintain sustainability on noise scheduling.
  \item Elimination of noise levels between $x_T$ and $x_0$ on the curriculum which relies on sinusoidal function.
\end{itemize}

In this study, it is proven that high level noises on noise scheduling has a crucial role to improve sampling quality and that is strengthened with a noise scheduling providing high variety of noise levels including high level noises on mini-batches. While polynomial function determines the index values for noise levels, noise levels are chosen from our predefined noise array $\sigma_{s_1}$ which is created before training. This approach is further reinforced by an innovative curriculum that relies on a sinusoidal function. It aims to diminish noise levels between $x_T$ and $x_0$ by choosing from the primary noise array $\sigma_{s_1}$  in a linear manner.

\section{Experimental Design}

\subsection{High Noise Level Experiments and Polynomial Noise Scheduling}

In this section, we evaluate the experimental results conducted within the scope of adding extra high noise levels manually to scheduled noises with a log-normal distribution. As it is claimed 'High Noises is a Must', the experiments begin by adding high-noise levels gradually on the mini-batches based on percentage ratio of the mini-batch size. This experiments proves that necessities of a noise scheduling comprising high-level noises beside high weighted low-level noises in a noise scheduling.

The high noise levels ranging between $40$ and $80$ are added randomly on mini-batches, with the length starting from $4$ to $12$ percent of the length of mini-batches. To represent the effect of high-noise levels on a mini-batch, the model employed in this experimental section is determined with $2$ residual blocks. 

The experiments reveals that adding minor weighted high noise levels on mini-batches increase denoising performance. As it is represented in table \ref{tab:table_high_noise}, while adding high level noise levels with lower percentage ratios can enhance denoising performance, adding high level noise at $10\%$ of the mini-batch length has effects on denoising performance conversely.

\begin{table*}[ht]
\scriptsize
\centering
\caption{Evaluation of denoising performance of models trained with noise schedules, including high noise levels proportional to the size of mini-batches.}
\resizebox{\linewidth}{!}{%
\begin{tabular}{l cccccccccc} 
\toprule
Model Versions  & Batch Size & Training Steps & Number of Res. Blocks     & High Level Noise Ratio  &N & Noise Scheduling & Attention Resolutions  & FID Score \\   
\midrule

Improved Model 1 & 1024 & 100000 & 2    & 0\% &100  & Log-Normal & [16,8]        & 66.51  \\ 
\midrule

Improved Model 2 & 1024 & 100000 & 2    & 2\% &100  & Log-Normal & [16,8]        & 75.34  \\ 
\midrule
Improved Model 3  & 1024 & 100000  & 2      & 3\% &100  & Log-Normal & [16,8]    & 43.81  \\  
\midrule
Improved Model 4  & 1024 & 100000  & 2      & 4\% &100  & Log-Normal & [16,8]     & 40.32  \\   
\bottomrule
Improved Model 5  & 1024 & 100000  & 2      & 5\% &100  & Log-Normal & [16,8]     & 73.24  \\   
\bottomrule
\end{tabular}%
}
\label{tab:table_high_noise}
\end{table*}
The results represented in Table \ref{tab:table_high_noise} demonstrate that implementing a log-normal noise distribution with high noise levels enhances denoising performance when the high-level noises reach $5\%$ of the total number of mini-batch size. Based on these results, we propose polynomial noise scheduling which represents high-weighted low noise levels and low-weighted high noise levels. Additionally, the degree of polynomial noise distribution can be determined by the user, depends on how much low and high level noises desired in noise scheduling.

\begin{equation}
    \sigma_{N(k)} =  \sigma_{s_1}[ \lfloor {i} * \frac{s_1}{N(k)} \rceil ], \text{where } i \in \llbracket 0, N(k) \rrbracket  
    \label{eq:choose_from_main}
\end{equation}

\begin{equation}
    \sigma_i =   \sigma_{N(k)}[ \lfloor (\frac{i}{d -1})^c * (N(k)-1)   + n(1,0)  \rceil]  , \text{where } i \in \llbracket 0, N(k)-1 \rrbracket  
    \label{eq:schedule_from_sigma_Nk}
\end{equation}
The polynomial noise scheduling function $\sigma_i$, as it represented in equation \ref{eq:schedule_from_sigma_Nk}, includes a $c$ parameters which provides letting the user to adjust polynomial curve. That presents a flexibility to determine the weight of low-level and high-level noises added on the mini-batches. Additionally, polynomial noise scheduling is augmented with a normal distribution which is added on mini-batches to introduce randomness in the noise distribution.

The noise levels added to mini-batches, denoted as $\sigma_i$, are chosen from $\sigma_{N(k)}$, which is generated simply based on the discretization steps $N(k)$.  The predefined noise array $\sigma_{s_1}$ is further explained in the section titled 'Analysis of Karras Noise Generation and Discretization Steps'.

The noise distributions for both approach, it is clear that proposed approach includes high variety of noise levels and also conserves the high weight of low level noises on the distribution. That ensures the model learning to high variety of noise levels conversely to the improved technique.
  
\subsection{Analysis of Karras Noise Generation and Predefined Noise Levels }

Karras noise scheduling is introduced as a robust algorithm with resulting analyzing sampling trajectories and their discretizations euclidian distance \cite{Karras}. Although this approach outperforms with consistency model training techniques, it is important to analyze and find the optimum utilizing technique. 

The implementation of Karras noise scheduling in consistency model training techniques generally is grounded on generation of new noises scheduling respect to the discretization steps changes. This utilization has a major drawback which is related to differentation of noises even the discretization steps are closer as $1$ step. To address this drawback, it is necessary to analyze the sensitivity of deep neural networks under various input conditions \cite{DNNanalysis}.

Based on this, we propose a simple approach which removes the generation of Karras noise schedules when discretization steps are changed. Instead of generation of Karras noise schedules for each discretization steps changes, a predefined Karras noise vector $\sigma_{s_1}$ with the length of $s_1$ is created before training phase, that prevents the generation of unique noise levels whenever discretization steps are changed. It makes possible to gather appropriate noise levels according to the discretization steps. This approach has identical method such as represented in Karras noise generation algorithm, which differs $N(k)$ determined as $s_1$. Noise generation approach for the predefined noise array is represented in equation \ref{eq:main_noise_array}.

\begin{equation}
    \sigma_{s_1} = \left(\sigma_{\min}^{1/\rho} + \frac{i-1}{s_1 -1} \cdot (\sigma_{\min}^{1/\rho} - \sigma_{\min}^{1/\rho})\right)^\rho, \text{where } i \in \llbracket 0, s_1 \rrbracket , \rho=7 , \sigma_{min}=0.002,  \sigma_{max}=80
    \label{eq:main_noise_array}
\end{equation}
By eliminating Karras noise scheduling for each change in discretization steps, the model can maintain consistent noise levels even when the discretization steps are adjusted to closer values.

\begin{equation}
\sigma_{N} =
\begin{cases}
\sigma_{s_1}[0,1,... s_1 ] & \text{if } N=s_1 \\
\sigma_{s_1}[0,2,4,... s_1 ] & \text{if } N(k) \leq \frac{s_1}{2} \\
\sigma_{s_1}[0,4,8,... s_1 ] & \text{if } N(k) \leq \frac{s_1}{4}
\end{cases}
\label{eq:discretization_from_main}
\end{equation}

To clarify this, equation \ref{eq:discretization_from_main} provides a basic simulation for scenarios where $N(k)$ equals different fractions of $s_1$. The highest advantage of this method, the number of unique noise levels never exceed the higher discretization steps, conversely to the improved noise distribution.

\subsection{Analysis of Curriculum and Elimination of Learned Noise Levels by Sinusoidal Curriculum  }

\begin{figure}[ht]
\centering
\includegraphics[width=\linewidth]{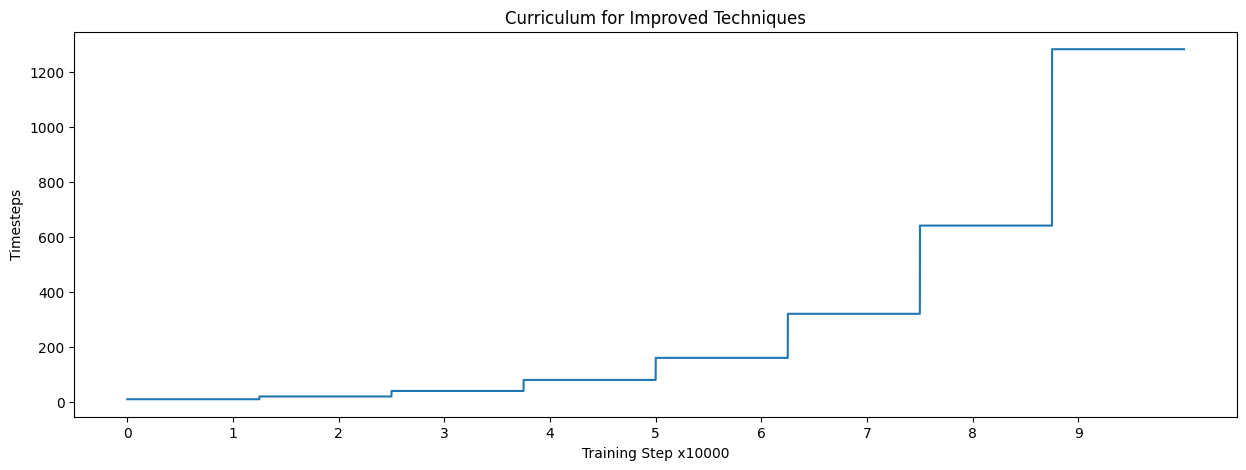}
\caption{The curriculum for Improved Training Techniques for Consistency Models \cite{Song2023-imp}. ($s_0=20, s_1= 1280$) }

\label{fig_imp_curriculum}
\end{figure}

The curriculum, which dictates the generation of noise steps during each training iteration, plays a pivotal role. Understanding the significance of the curriculum necessitates acknowledging that while $\xi$ learns denoising from $x_t$ to $x_0$, $\theta$ converges and learns denoising from $x_{t+1}$ to $x_t$.This implies that even slight increases or decreases in the number of discretization steps lead to improved performance. In Improved Techniques, the curriculum doubles itself at each specific training steps which that results a significant increase in number of discretization steps as it is represented in figure \ref{fig_imp_curriculum}.  These sudden and notable changes in number of discretization steps require the model to adapt to the new noise levels over a longer period of time.

To address this issue, we propose to utilize a technique based on a sinusoidal function, which facilitates smooth increases and decreases during the training steps transition. This enhances the model's adaptability to the new noise levels. In proposed approach, $s_0$ and $s_1$ are adjusted as $20$ and $250$ respectively.

\begin{figure}[ht]
\centering
\includegraphics[width=\linewidth]{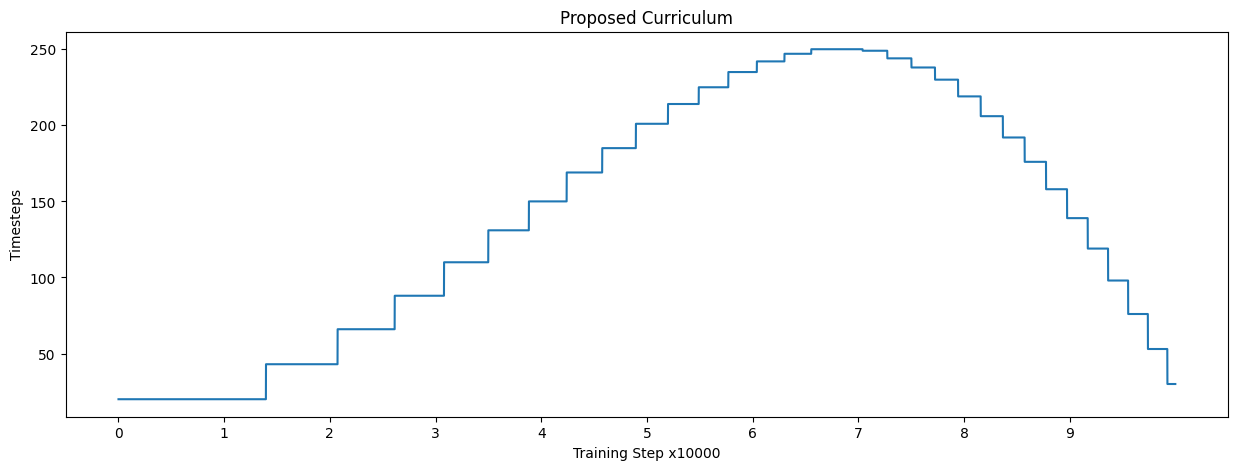}
\caption{The proposed curriculum relies on sinusoidal function. ($s_0=20, s_1= 250$) }
\label{fig_sinus_curriculum}
\end{figure} 

The underlying reason for employing a sinusoidal function is to make sure the model is capable of learning various time steps at the highest number of discretization steps. After reaching the highest number of discretization steps, the curriculum function starts to decrease the number of discretization steps in order to strengthen the learning of denoising from $x_T$ to $x_0$. With the help of advancement of predefined noise levels, the noise levels included by $\sigma_i$ are eliminated throughout the end of training.

\begin{equation}
    N(k) = \lceil |  (s_1 - s_0) * sin( \frac{\lfloor \frac{k^{\rho/4}}{K^{\rho/4}} * \pi \rfloor * 10 }{10} ) + s_0 | \rceil,  k \in \llbracket 0, K \rrbracket, \text{where K is the total training iteration}
    \label{eq:curriculum}
\end{equation}

In our approach, the maximum number of discretization steps, denoted as $s_1$, for the curriculum is adjusted to $250$ to mitigate abrupt increases in the curriculum during training. The curriculum utilized in our approach is represented in equation \ref{eq:curriculum}. $N$ represents the number of discretization steps or the number of noise levels are generated at the current time step which is denoted as $t_s$. Other constants, such as $s_0$ and $s_1$, represent the final and initial time steps, which are determined as $20$ and $250$, respectively. The number of training step, $T$, is adjustable as desired and $t$ represents current training step which is able to increase until $T$. $\rho$ variable has same value as it is used in the curriculum. The proposed curriculum also is depicted in figure \ref{fig_sinus_curriculum}.

\begin{figure}[ht]
\centering
\includegraphics[width=0.7\linewidth]{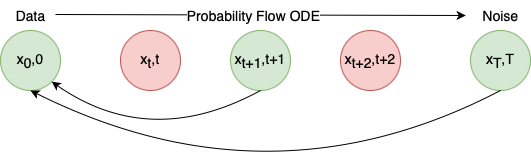}

\caption{Representation for elimination of learned  noise levels $x_t$ from curriculum during training  }

\label{fig_curriculum_elimination}
\end{figure}

\section{Experiments and Results}

\subsection{Experimental Environment and Dataset}
The experimental environment consists of two NVIDIA A6000 graphics cards. All model versions are trained with the same attention resolution and number of residual blocks, which are assigned as $[8,16]$ and $8$, respectively. The main data set chosen for training is CIFAR-10, which comprises 32x32 images and a total of 50,000 samples \cite{cifar10}.

\subsection{Training Results} 
A series of experiments were conducted to understand the effect of different noise scheduling techniques based on various distributions. Additionally, we observe the efficiency of hyper parameters on the performance using these noise scheduling methods. The experimental results reveals that a noise scheduling including high weighted low noise levels with guaranteeing the highest noise level leverages FID results.

Our experiments consist of three steps. The first step involves maintaining the number of discretization steps constant to examine polynomial noise scheduling. To explore the optimal curve parameter, we conducted the first four experiments with a model containing 4 residual blocks, steadily increasing the polynomial curve as represented in the table \ref{tab:table_training_polynomial}. 

\begin{table*}[hbt!]
\scriptsize
\centering
\caption{Hyper-parameters and noise scheduling parameters for polynomial noise scheduling}
\resizebox{\linewidth}{!}{%
\begin{tabular}{l cccccccccc} 
\toprule
Model Versions  & Batch Size & Training Steps & Number of Res. Blocks     & Scale Shifting  &N & Noise Scheduling & Attention Resolutions & Curve & FID Score \\  
\midrule
Model P1  & 1024 & 100000 & 4     & False &20 & polynomial & [16,8]      & 2   & 38.28 \\ 
\midrule
Model P2  & 1024 & 100000 & 4    & False &20  & polynomial & [16,8]      & 3   & 34.05  \\ 
\midrule
\textbf{Model P3} & 1024 & 100000  & 4      & False &20  & polynomial & [16,8]      & 4   &  33.54\\ 
\midrule
Model P4 & 1024 & 100000  & 4     & False &20  & polynomial & [16,8]      & 5   & 37.94  \\  
\bottomrule
\end{tabular}%
}
\label{tab:table_training_polynomial}
\end{table*}

The table \ref{tab:table_training_polynomial} reveals that,  a polynomial function with power 4 has a close optimum balance between low-level and high-level noises are represented in a noise scheduling. In first experimental step, the $model$ $P3$ has the best FID results when compared the other models have different polynomial curve in noise distribution function. 
To show the efficiency of the sinusoidal curriculum on FID results, the hyperparameters of the best model remain constant throughout the execution of the following experiments. The results  gathered from the model utilizing the sinusoidal curriculum are compared with other models using improved curriculum. The results for comparison of improved curriculum and sinusoidal curriculum are represented in table \ref{tab:table_training_optimum}.

\begin{table*}[hbt!]
\scriptsize
\centering
\caption{Comparison of the sinusoidal and the models are trained with improved techniques}
\resizebox{\linewidth}{!}{%
\begin{tabular}{l cccccccccc} 
\toprule
Model Versions  & Batch Size & Training Steps & Number of Res. Blocks     & Scale Shifting  &N & Noise Scheduling & Attention Resolutions & Curve & FID Score \\  
\midrule
\textbf{Model P5}  & 1024 & 100000 & 4     & False &Dynamic & polynomial & [16,8]      & 3   & 30.48 \\ 
\midrule
CM Improved 1  & 1024 & 100000 & 4    & True &20  & Log-Normal & [16,8]      & 3   & 48.80  \\ 
\midrule
CM Improved 2  & 1024 & 100000  & 4      & False &Dynamic  & Log-Normal & [16,8]      & 3   & 49.19  \\  
\bottomrule
\end{tabular}%
}
\label{tab:table_training_optimum}
\end{table*}
The sinusoidal curriculum takes  advantage of eliminating the noisy steps $x_t$ selected from predefined noise array between $x_0$ and $x_T$. As it is represented in table \ref{tab:table_training_optimum}, the $model$ $P5$ proves that model is able to learn each noise level after the curriculum reached out the maximum discretization steps. 
\iffalse
REMOVE RAYLEIGH ? 
The second step involves increasing the number of residual blocks to demonstrate the efficacy of polynomial noise scheduling with the model that produced the lowest FID results gathered from the first step of the experiments. Referring to Table \ref{tab:table_training_polynomial}, the model $model$ $P2$ generates the highest quality samples and the lowest FID results with residual blocks $4$. After determining the optimal parameter for the polynomial curve, $model P2$ is further evaluated with different hyperparameters as detailed in Table \ref{tab:table_training_optimum}.  In the final step, we employ our sinusoidal curriculum approach with the best model obtained from the second step, along with its corresponding hyperparameters. 
\begin{table*}[hbt!]
\scriptsize
\centering
\caption{Hyper-parameters and noise scheduling parameters for Rayleigh noise scheduling}
\resizebox{\linewidth}{!}{%
\begin{tabular}{l cccccccccc} 
\toprule
Model Versions  & Batch Size & Training Steps & Number of Res. Blocks     & Scale Shifting  &N & Noise Scheduling & Scale  & FID Score \\ 
\midrule
Model R1 & 256 & 100000  & 6      & False &20  & Rayleigh & 2         & 59.43  \\ 
\midrule
Model R2 & 256 & 100000  & 8      & True &20   & Rayleigh & 2          & 49.1   \\ 
\midrule
Model R3 & 256 & 100000  & 8    & False &20  & Rayleigh & 3 & 73.07  \\ 
\midrule
Model R4  & 256 & 100000 & 8     & False &20  & Rayleigh & 4         & 86.79  \\   
\bottomrule
\end{tabular}%
}
\label{tab:table_training_rayleigh}
\end{table*}
\fi
\subsection{Experimenal Results} 

Training the models with various hyperparameters reveals that the number of training steps, 100.000, is insufficient for the model to learn the noise variations occurring over short periods of time. To handle with short training time and increasing the capability of model learning, the sinusoidal curriculum takes the advantage of decreasing the number of discretization steps. As explained, the model learns to close the gap between $x_t$ and $x_0$ as the number of discretization steps decreases.

The model trained with improved technique and the other trained with our proposed technique are compared for different inference steps from $1$ to $4$. Despite assigning the same hyperparameters to  all trained models in our experiments, our proposed approach yields a lower FID score compared to what is reported in the paper for improved techniques. Also, it should be known that FID scores for the models trained improved technique output a close FID result as it is reported by \cite{Song2023-imp}, regarding the number of residual employed blocks.

\section{Conclusion}
Our enhancements to noise scheduling and curriculum address the need for a balanced noise distribution and controlled progression of noise steps within the curriculum. We examined the impact of the high noise levels in a noise distribution comprising the high-weighted low noise levels. By pre definition of Karras noises, it is prevented that the unique noise steps resulting from Karras noise generation algorithm. This approach is supported by our sinusoidal curriculum, which eliminates learned noise steps after the highest number of discretization steps throughout the learning process. All examinations and comparisons are conducted with the same U-Net architecture and hyper-parameters to show efficiency of our approach. Remarkably, our fundamental changes leverage FID results when compared to the latest improvements made on consistency models.

\section{Acknowledgements}

The project described was supported by the NIH National Center for Advancing Translational Sciences through grant number UL1TR001998.  The content is solely the responsibility of the authors and does not necessarily represent the official views of the NIH.

\begin{figure}[ht]
\centering
\includegraphics[width=0.7\textwidth]{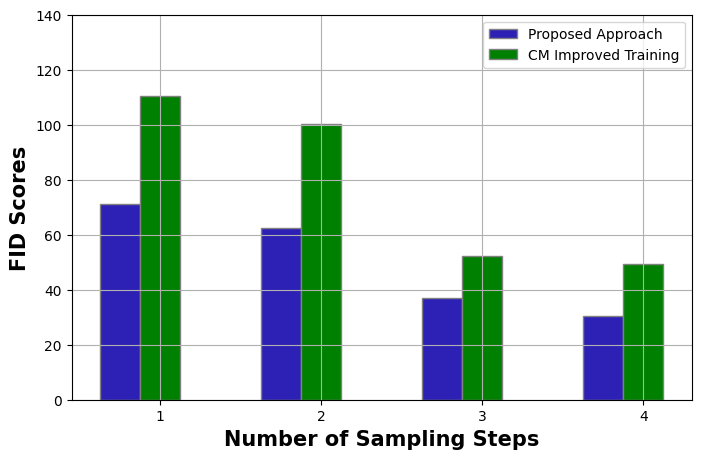}
\caption{The comparison of improved training and our proposed training technique }
\label{fig_fid_compare}
\end{figure}

\begin{figure}[ht]
\centering
\includegraphics {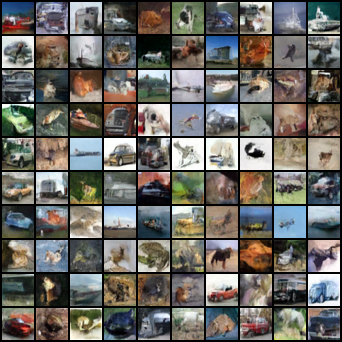}
\caption{Four-step samples from the model trained with our technique on Cifar10 32x32 (FID= 30.48) }
\label{fig_sample_1}
\end{figure}

\begin{figure}[ht]
\centering
\includegraphics {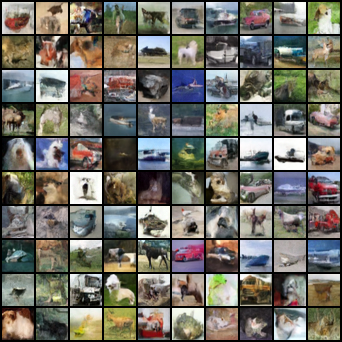}
\caption{Four-step samples from the model trained with our technique on Cifar10 32x32 (FID= 30.48) }
\label{fig_sample_2}
\end{figure}

\begin{figure}[ht]
\centering
\includegraphics {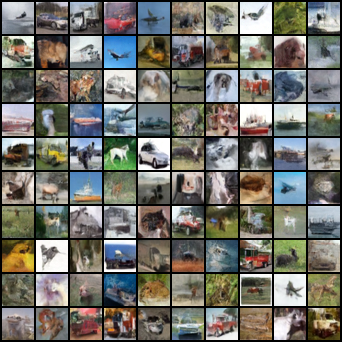}
\caption{Four-step samples from the model trained with our technique on Cifar10 32x32 (FID= 30.48) }
\label{fig_sample_3}
\end{figure}

\bibliographystyle{unsrtnat}
\bibliography{references}  %%% Uncomment this line and comment out the ``thebibliography'' section below to use the external .bib file (using bibtex) .

\begin{thebibliography}{6}
\providecommand{\natexlab}[1]{#1}
\providecommand{\url}[1]{\texttt{#1}}
\expandafter\ifx\csname urlstyle\endcsname\relax
  \providecommand{\doi}[1]{doi: #1}\else
  \providecommand{\doi}{doi: \begingroup \urlstyle{rm}\Url}\fi

\bibitem[Song et~al.(2020)Song, Sohl-Dickstein, Kingma, Kumar, Ermon, and Poole]{Song2020-sb}
Yang Song, Jascha Sohl-Dickstein, Diederik~P Kingma, Abhishek Kumar, Stefano Ermon, and Ben Poole.
\newblock Score-based generative modeling through stochastic differential equations.
\newblock 2020.

\bibitem[Song et~al.(2023)Song, Dhariwal, Chen, and Sutskever]{Song2023-cm}
Yang Song, Prafulla Dhariwal, Mark Chen, and Ilya Sutskever.
\newblock Consistency models.
\newblock 2023.

\bibitem[Song and Dhariwal(2023)]{Song2023-imp}
Yang Song and Prafulla Dhariwal.
\newblock Improved techniques for training consistency models.
\newblock 2023.

\bibitem[Karras et~al.(2022)Karras, Aittala, Aila, and Laine]{Karras}
Tero Karras, Miika Aittala, Timo Aila, and Samuli Laine.
\newblock Elucidating the design space of diffusion-based generative models.
\newblock 2022.

\bibitem[Montavon et~al.(2018)Montavon, Samek, and M\"{u}ller]{DNNanalysis}
Grégoire Montavon, Wojciech Samek, and Klaus-Robert M\"{u}ller.
\newblock Methods for interpreting and understanding deep neural networks.
\newblock \emph{Digital Signal Processing}, 73:\penalty0 1–15, February 2018.
\newblock ISSN 1051-2004.
\newblock \doi{10.1016/j.dsp.2017.10.011}.
\newblock URL \url{http://dx.doi.org/10.1016/j.dsp.2017.10.011}.

\bibitem[Krizhevsky(2009)]{cifar10}
Alex Krizhevsky.
\newblock Learning multiple layers of features from tiny images.
\newblock pages 32--33, 2009.
\newblock URL \url{https://www.cs.toronto.edu/~kriz/learning-features-2009-TR.pdf}.

\end{thebibliography}





% \end{thebibliography}

\end{document}